\documentclass[letterpaper, 10 pt, conference]{ieeeconf} 
\IEEEoverridecommandlockouts                               
\usepackage{graphicx}
\usepackage{amsmath}
\usepackage{xcolor}
\usepackage{amssymb}
\usepackage{float}
\usepackage{subfigure}
\let\oldemptyset\emptyset
\usepackage{setspace}
\usepackage{comment}
\usepackage{color}

\renewcommand{\baselinestretch}{0.965}

\title{\LARGE \bf
A Control Architecture for Provably-Correct Autonomous Driving}

\author{Erfan Aasi$^{1}$, Cristian Ioan Vasile$^{2}$ and Calin Belta$^{1}$

\thanks{*This work was partially supported by the NSF under grant IIS-1723995 and IIS-2024606 at Boston University.
}
\thanks{$^{1}$Erfan Aasi and Calin Belta are with Department of Mechanical Engineering,
        Boston University, 
        Boston, MA 02215, USA
        {\tt\small eaasi@bu.edu},{\tt\small cbelta@bu.edu}}
\thanks{$^{2}$Cristian Ioan Vasile is with the Mechanical Engineering and Mechanics Department, Lehigh University,
        Bethlehem, PA 18015, USA
        {\tt\small cvasile@lehigh.edu}}
}

\begin{document}

\maketitle
\thispagestyle{empty}
\pagestyle{empty}

\begin{abstract}

This paper presents a novel two-level control architecture for a fully autonomous vehicle in a deterministic environment, which can handle traffic rules as specifications and low-level vehicle control with real-time performance.
At the top level, we use a simple representation of the environment and vehicle dynamics to formulate a linear Model Predictive Control (MPC) problem. We describe the traffic rules and safety constraints using Signal Temporal Logic (STL) formulas, which are mapped to mixed integer-linear constraints in the optimization problem. The solution obtained at the top level is used at the bottom-level to determine the best control command for satisfying the constraints in a more detailed framework. At the bottom-level, specification-based runtime monitoring techniques, together with detailed representations of the environment and vehicle dynamics, are used to compensate for the mismatch between the simple models used in the MPC and the real complex models. We obtain substantial improvements over existing approaches in the literature in the sense of runtime performance and we validate the effectiveness of our proposed control approach in the simulator CARLA.

\end{abstract}

\section{INTRODUCTION}
\label{section:introduction}
Recent advances in computational software and electronic devices have led to development of several automated safety features for modern vehicles over the last few years, e.g., Adaptive Cruise Control \cite{vahidi2003research}, and lane keeping systems \cite{liu2007development}. However, the dynamically changing nature of urban driving environments, and necessity of obeying a diverse set of traffic rules, limit the performance of these safety systems to relatively low-complexity driving scenarios.

To overcome this problem, a variety of control and planning algorithms have been proposed, some of which are based on Model Predictive Control (MPC) \cite{camacho2013model}. In MPC, models of the autonomous vehicle (referred to as {\em ego}), traffic and environment at the current state are used to predict the future behavior of the traffic participants over the MPC's finite horizon. The future-prediction property, together with the capability of systematically handling constraints, have made MPC a popular control approach for self-driving cars \cite{zhang2016model}, \cite{anderson2010optimal}, \cite{faulwasser2009model} \cite{weiskircher2017predictive}.
In \cite{schwarting2017parallel} and \cite{erlien2015shared}, the authors formulated MPC problems for providing safe motions, with the goal of minimizing the required control intervention in the human driver's inputs. Uncertainty in the driver's behavior was investigated in the predictive control algorithms proposed by \cite{gray2013robust} and \cite{liu2015stochastic}. In \cite{turri2013linear}, the authors presented predictive control methods for path following on slippery roads based on linear time varying dynamic models. 
These approaches are computationally expensive. Any increase in the complexity of the models (ego's dynamics, traffic, environment, uncertainty) leads to an explosion in the runtime. As a result, these approaches cannot handle real traffic scenarios. 

There has been a growing trend recently to account for traffic complexities involving human drivers \cite{sadigh2016planning}, \cite{schwarting2019social}, and rules of the road \cite{karlsson2018multi}, \cite{tumova2013least}, \cite{vasile2017minimum}, \cite{censi2019liability}. For the later, temporal logics \cite{baier2008principles} were proposed to formalize the rules of the road. MPC problems with Linear Temporal Logic (LTL) constraints have been investigated in \cite{ding2014ltl}, \cite{wongpiromsarn2012receding}. In \cite{raman2014model} and \cite{sadraddini2015robust} control strategies have been proposed for discrete-time systems subject to Signal Temporal Logic (STL) formulas \cite{donze2010robust}, which were encoded as Mixed-Integer Linear Constraints (MILCs) in an MPC framework. Even though such problems can be solved relatively fast, existing approaches generally do not satisfy real-time performance requirements.

\begin{figure}[htb]
    \centering
    \subfigure[High-level controller]
    {\includegraphics[width=0.60\columnwidth]{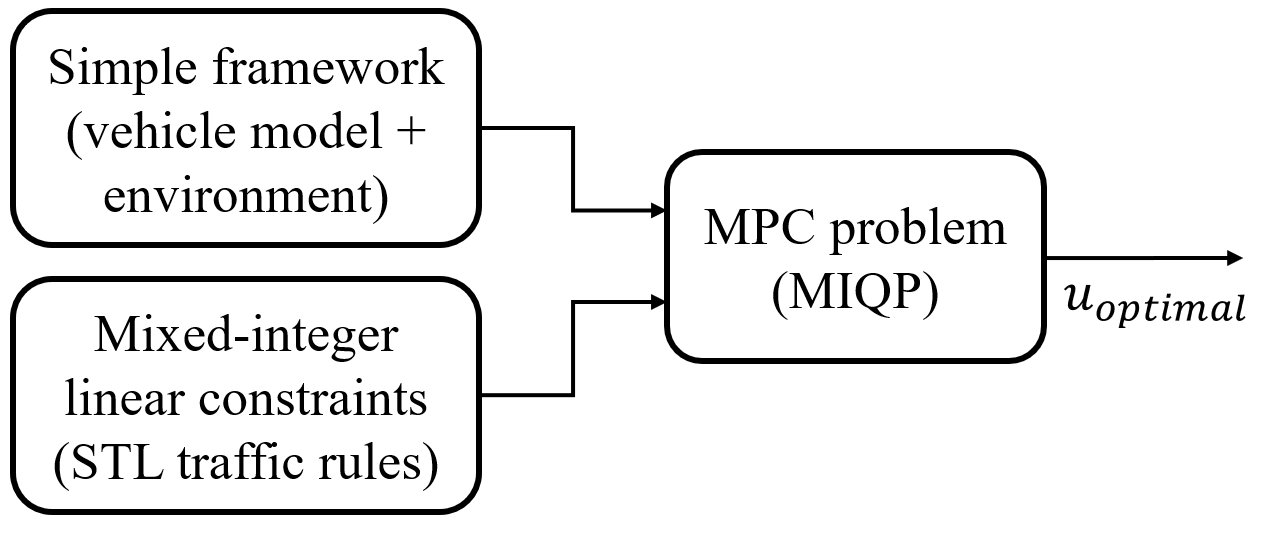}\label{fig:highchart}}
    \subfigure[Low-level controller]
    {\includegraphics[width=0.75\columnwidth]{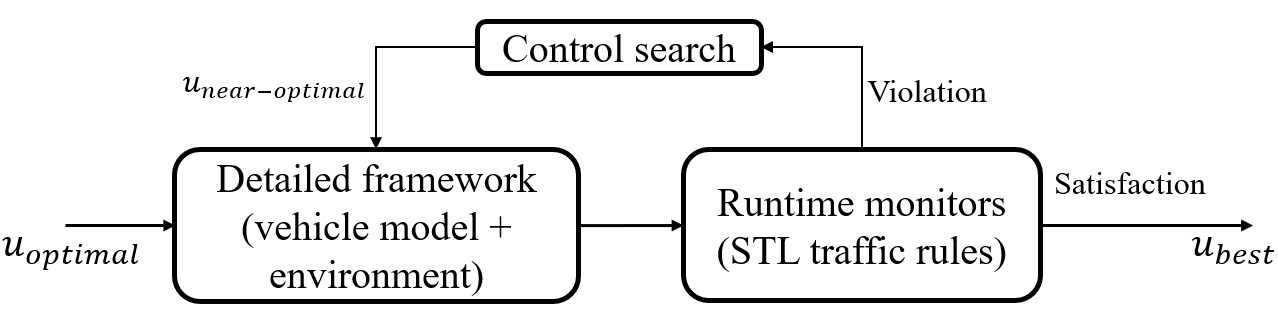}\label{fig:lowchart}} 
    \caption{Overview of the proposed method: (a) high-level controller based on MIQP, (b) low-level controller based on runtime monitoring}
    \label{fig:flowchart}
    \vspace{-6mm}
\end{figure}

In this paper, we consider rules of the roads expressed using STL formulas, together with collision avoidance requirements, and a cost function that penalizes, e.g., acceleration or heading changes. We use runtime monitoring \cite{bartocci2018specification}, \cite{deshmukh2017robust} to alleviate computational burden associated with control via complex dynamics, models, and constraints.
We specifically focus on meeting the runtime performance criteria. The problem we are tackling is difficult even in the deterministic regime, due to computational complexity arising from the large number of constraints. We propose a two-level control scheme, which solves the control problem at the top level with simple models and checks the correctness of the solutions at the bottom level against complex models (see Fig.~\ref{fig:flowchart}). 

At the top level, based on a successive on-line linearization of a simple bicycle model and a simple representation of the environment, an MPC framework with a quadratic objective function is formulated. At this level, the STL specifications are encoded as mixed-integer linear constraints and imposed in the optimization problem, which leads to a Mixed-Integer Quadratic Programming (MIQP) problem. The obtained optimal controller is used in the bottom level controller to find the best control input for satisfying the traffic rules in a more detailed framework, by applying runtime monitoring on the fly over the trajectories of ego.
Through computational experiments performed in the urban traffic simulator CARLA, we show significant performance benefits of the proposed method, in terms of runtime and solution quality, against existing NMPC approaches.

\section{Problem Statement}
\label{section:problemstatement}
We consider discrete-time continuous-space models for the dynamics of the vehicles in the environment. We assume uniform discretization intervals $\Delta t$ (each time step $t$ is an integer multiple of $\Delta t$). Ego is required to follow a reference path denoted by $\mathcal{R}$ between an initial and a final point. The MPC's finite horizon consists of $H$ time steps, and the total finite number of time steps of the vehicle's operation is denoted by $T$, which is unknown. All of the vector variables are shown in bold and the Euclidean norm of a vector $\mathbf{x}$ is denoted by $|| \mathbf{x} ||$. 

\subsubsection{Dynamic Model of Ego}
\label{section:generaldynamicmodel}
 The state vector of ego consists of the position of the vehicle $(X_t,Y_t)$ in the absolute coordinate frame of the environment, its heading (yaw angle) $\psi_t$ and its velocity vector $(\dot{X}_t,\dot{Y}_t)$ of magnitude $v_t =\|(\dot{X}_t, \dot{Y}_t)\|$. The control input is denoted by $\mathbf{u}_t = [\delta_t, \gamma_t]^\top$, where $\delta_t \in [\delta_{min}, \delta_{max}]$ is the steering angle and $\gamma_t \in [-1,1]$ is the standard throttle ($\gamma > 0$) and braking ($\gamma < 0$) input.
The model of ego is given by 
\begin{equation}
\label{eqn:generaldynamic}
    \boldsymbol{\zeta}_{t+1} = f (\boldsymbol{\zeta}_t, \mathbf{u}_t), 
\end{equation}
where $f$ is specified in Sections~\ref{section:simplebicycle} and~\ref{section:detailedbicycle}.

\subsubsection{Model of Environment}
\label{section:generalenvironmentmodel}
The environment consists of lanes, traffic signs, and traffic participants such as vehicles, pedestrians and bicyclists. For simplicity, we assume that all traffic participants are vehicles, but our method is applicable to general urban driving environments with any type of traffic participants. We assume the sensors and cameras of ego detect and provide necessary information about traffic signs and other vehicles in a specific nearby radius, denoted by $r_{near}$
(details in Sections~\ref{section:simpleenvironment} and~\ref{section:detailedenvironment}).

\subsubsection{Reference Path}
We assume a reference path $\mathcal{R}$ is provided to ego, which is an untimed finite sequence of $M$ points $P_i$: $\mathcal{R} = \{P_1, ..., P_M\}$, where each $P_i$ specifies a position $(X_{r,i}$, $Y_{r,i})$ and a heading $\psi_{r,i}$. We assume the positions are in the middle of the lanes and they are equally spaced from each other (see Fig.~\ref{fig:desiredtrajectory}). 


\subsubsection{Objective Function}
\label{section:objectivefunction}
Ego is required to follow the reference path as closely as possible. The objective function $\mathcal{F}_{obj}(.)$ at each time step accounts for ego's control inputs, the comfort of the driver determined by changes in the inputs, and tracking error with respect to the reference path.
At each time step $t$, given the state of ego $\boldsymbol{\zeta}_t$ and the reference path $\mathcal{R}$, we consider the trajectory error term as $e_t = \mathcal{E}(\mathcal{R}, \boldsymbol{\zeta}_t)$, where the function $\mathcal{E}$ computes the Euclidean distance between current position of ego and the nearest position on the path $\mathcal{R}$. Thus, we formulate the objective function as
\begin{equation}
\label{eqn:objectivefunction}
    \mathcal{F}_{obj}(\mathbf{u}_t, \mathbf{u}_{t-1}, e_t) = \|\mathbf{u}_t\|^2 + \|\mathbf{u}_t - \mathbf{u}_{t-1}\|^2 + e_t^2 .
\end{equation}

\subsubsection{Traffic Rules}
\label{section:trafficrules}
In this paper, we capture traffic rules using STL formulae~\cite{maler2004monitoring} over predicates $p$ in the state $\boldsymbol{\zeta}_t$ of system  (\ref{eqn:generaldynamic}). Informally, STL formulas $\phi$ are formed using Boolean connectives, such as $\lnot$ $\land$ $\lor$, and temporal operators, such as  $\mathcal{U}_{[a,b]}$ (until), $\square_{[a,b]}$ (always) and $\lozenge_{[a,b]}$ (eventually). The semantics of STL formulas is interpreted over state trajectories $\boldsymbol{\zeta}_t$. We use $\boldsymbol{\zeta}_t \models \phi$ to denote that $\boldsymbol{\zeta}_t$ satisfies $\phi$ at time $t$. For example, $\boldsymbol{\zeta}_t \models \square_{[a,b]}\phi$ means that $\phi$ is satisfied for all times in the interval $[a,b]$ along $\boldsymbol{\zeta}_t$; $\boldsymbol{\zeta}_t \models \lozenge_{[a,b]}\phi$ means that $\phi$ is satisfied for some time in the interval $[a,b]$. 

We consider collision avoidance as a traffic rule. Moreover, we do not explicitly require the vehicle to stay in lane - we assume this constraint is satisfied as the vehicle follows the reference path, which is made of points in the middle of lanes. We denote the finite set of STL traffic rules by $\boldsymbol{\Phi}$. We use $\boldsymbol{\Phi}_{active, t} \subseteq \boldsymbol{\Phi}$ to denote the subset of active traffic rules at time step $t$. 


Consider for example the collision avoidance constraint, where ego is required to avoid collisions with the other vehicles in the environment at all times. This traffic rule can be formulated as $\square_{[0, H]} \textit{ (collision\_avoidance)}$, where $(collision\_avoidance)$ indicates that ego does not collide with any other vehicle, over the MPC's horizon $H$.
In Sections~\ref{section:simpleenvironment} and~\ref{section:detailedenvironment} we explain how we check value of this predicate in each level of the proposed controller.

\subsubsection{\textbf{Problem Formulation}}
\label{section:problemformulation}

Given the environment of ego, its dynamic model~\eqref{eqn:generaldynamic}, the reference path $\mathcal{R}$, and the set of traffic rules $\boldsymbol{\Phi}$, find a control sequence $\mathbf{u}^*_{1:T}$ such that:
\begin{align}
\label{eqn:problemformulation}
    \mathbf{u}^*_{1:T} & = \underset{\mathbf{u}_{1: T}}{\arg \min} \sum_{k=1}^{T} \mathcal{F}_{obj}(\mathbf{u}_k, \mathbf{u}_{k-1}, e_k) \\ \nonumber
    \text{s.t.} \quad \boldsymbol{\zeta}_{k+1} &= f(\boldsymbol{\zeta}_k, \boldsymbol{u}_k), \quad e_k = \mathcal{E}(\mathcal{R}, \boldsymbol{\zeta}_k), \quad \boldsymbol{\zeta}_k \models \phi,\\ \nonumber
    & \forall \phi \in \boldsymbol{\Phi}_{active,k} , \quad k = 1,...,T
\end{align}


At each time step $k$ we need the total computation time of updating the control command to be less than the time step $\Delta t$ between discrete times $k$ and $k+1$.
Otherwise, the delays in applying the control inputs can lead to instability of the system and late reaction to outside events (see e.g., \cite{baillieul2007control}). 
In this paper, we require the solution of~\eqref{eqn:problemformulation} to be computed in less than $\Delta t=0.1$ seconds ($10\mathit{Hz}$). This rate is a common requirement for low-speed urban driving.

\section{Solution}
\label{section:problemsolution}
We propose a two-level controller for driving the autonomous vehicle from its initial state to the goal (Fig.~\ref{fig:flowchart}) and satisfying the runtime performance requirement. At the high level we formulate a MPC problem, which uses a simple representation of ego's dynamic model and its environment to compute the optimal control input at each time step. At this level, the set of traffic rules that ego must satisfy are imposed as mixed-integer linear constraints to the MPC. The optimal control from the high-level controller is the used in the low-level to forward simulate the trajectory ego in a complex model of ego and the environment. STL-based runtime monitoring is used in this level to verify the correctness of ego's behavior. In the following, we detail each layer of the controller.
\subsection{\textbf{High-level Controller}}
In this level we use a simple bicycle dynamic model for ego and simple representations for traffic participants, to formulate a MPC problem with planning horizon $H$ and with a quadratic cost function $\mathcal{J}$. Traffic rules are translated to MILCs and the active rules at each time step are imposed to MIQP optimization problem.

\subsubsection{Simplified Vehicle Model}
\label{section:simplebicycle}
We consider a kinematic bicycle model as a simplified model for ego:
\begin{equation} \label{eqn:simplebicycle}
    [\dot{X}, \dot{Y}, \dot{\psi}, \dot{v}]^\top = [v \cos{\psi}, \, v \sin{\psi}, \, \frac{v}{l_f} \tan{\delta}, \, a]^\top
\end{equation}
where $l_f$ is the distance from center of the mass to the front axle, and $a$ is the acceleration (or deceleration) of the car. Generally there is a complicated relationship between the acceleration (or deceleration) of a vehicle and the applied throttle (or braking) $\gamma$. Here we assume there is an (approximate) map to convert between them and in the simulation results in Section  ~\ref{section:results}, we estimate this mapping by benchmark data. In (\ref{eqn:simplebicycle}) we have assumed only the front wheel's steering $\delta_f$ affects the orientation of the car and for simplicity we have shown $\delta_f$ by $\delta$. If we denote the state vector of simplified vehicle model at time step $t$ by $\boldsymbol{\zeta}_{t}^s = [X_t,Y_t,\psi_t,v_t]^\top$ and the input vector by $\mathbf{u}_t = [\delta_t,\gamma_t]^\top$, the time-discrete, linearized version of (\ref{eqn:simplebicycle}) is $\boldsymbol{\zeta}_{t+1}^s = f^s (\boldsymbol{\zeta}_{t}^s, \mathbf{u}_t)$.


\subsubsection{Simplified Environment Model}
\label{section:simpleenvironment}
Other vehicles in the environment are indexed by $i \in I = \{1,...,N\}$, and the position of the $i^{th}$ vehicle at the time step $t$ is denoted by $(X_{i,t}, Y_{i,t})$. We assume at each time step $t$, the sensors of ego are able to estimate the velocity and heading of the nearby vehicles, to predict their future trajectories over the period $[t, t+H]$. We use simple bicycle model in~\eqref{eqn:simplebicycle} to predict the trajectories of other vehicles assuming constant velocity and heading over the horizon $H$. There are no reference paths for other vehicles. However, we assume that they obey traffic rules and perform simple collision avoidance.

The set of vehicles closer than $r_{near}$ to ego at time step $t$ is denoted by $I_{near,t} \subset I$, i.e., $I_{near, t} = \{ i \in I \mid \| (X_t, Y_t) - (X_{i,t}, Y_{i,t}) \| \leq r_{near} \}$. To simplify ego's environment representation, we consider all vehicles in the environment as point-masses. To avoid collisions, the Euclidean distance between ego and nearby vehicles must be bigger than a safe distance $D_{safe}$. However, quadratic non-convex constraints are difficult to impose and lead to higher runtime for the algorithm. Therefore, we approximate the Euclidean norm with the 1-norm, and define the $(collision\_avoidance)$ predicate in the collision avoidance rule by $\forall i \in I_{near, t}: |X_t - X_{i,t}| + |Y_t - Y_{i,t}| \geq D_{safe}$. This requirement ensures collision avoidance, as the 1-norm upper bounds the Euclidean norm, and it can be imposed as a linear constraint to the optimization problem.
\subsubsection{Rules of the Roads}
Given the set of active traffic constraints $\boldsymbol{\Phi}_{active,t}$ at time step $t$, we use the techniques from \cite{raman2014model}, \cite{sadraddini2015robust} to translate the STL specifications to MILCs and impose them in the MPC problem over the time horizon $H$. If any of active STL specifications has a time horizon longer than the MPC's planning horizon, we keep track and adjust the involved signals in that specification to have a correct notion of receding horizon control for that specification. 
\subsubsection{MPC Formulation}
\begin{figure}[htb]
\centering
\includegraphics[width=0.45\columnwidth]{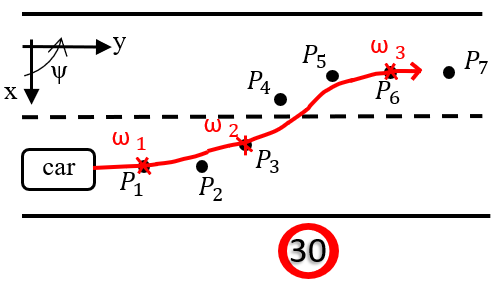}%
\vspace{-3mm}
\caption{Representation of reference path and desired trajectory. Following the traffic rules, the desired speed for the waypoint $\boldsymbol{\omega}_2$ and after will be considered $30 \, km/h$.}
\label{fig:desiredtrajectory}
\end{figure}

At each time step $t$, based on the current state and goal position of ego, a desired trajectory to follow over the period $[t, t+H]$ is extracted from the reference path, in the form of $H$ waypoints (see Fig.~\ref{fig:desiredtrajectory}). In more detail, each of the waypoints consists of the desired position $(X_{des,t}, Y_{des, t})$ and desired heading $\psi_{des, t}$ matching one of the points $P_i$ in the reference path, together with desired speed $v_{des, t}$. Based on the sensor outputs of ego, the desired speed is adjusted with respect to the situation of ego at each time step. Note that in the desired trajectory of the MPC framework, the dynamic model of ego is not considered, which means the desired trajectory may not be dynamically feasible to follow at all time steps of the MPC horizon. We denote the desired waypoint at time step $t$ by $\boldsymbol{\omega}_t = [X_{des,t}, Y_{des,t}, \psi_{des,t}, v_{des,t}]$, and the desired trajectory to follow at time step $t$ by the sequence $[\boldsymbol{\omega}_t, \boldsymbol{\omega}_{t+1}, ..., \boldsymbol{\omega}_{t+H}]$.

We consider a quadratic objective function at each time step of the MPC framework, composed of three terms: a control penalty, a step-to-step control change penalty, and the trajectory tracking error. Hence for the cost function $\mathcal{J}$ of the MPC problem we have:
\begin{equation}
\label{eqn:mpccost}
    \mathcal{J}(\boldsymbol{\zeta}^s_t, \textbf{u}_t, \textbf{u}_{t-1}) = \lVert \mathbf{u}_t \rVert^2 + \lVert \mathbf{u}_t - \mathbf{u}_{t-1} \rVert^2 + \lVert \boldsymbol{\zeta}_t^s - \boldsymbol{\omega}_t \rVert^2
\end{equation} 
And we can formulate the optimization problem as:
\begin{align}
\label{eqn:mpc}
    \mathbf{u^*}_{t:\;t+H-1} & = \underset{\mathbf{u}_{t: \; t+H-1}}{\arg\min} \sum_{k=t}^{t+H-1} \mathcal{J}(\boldsymbol{\zeta}^s_k, \mathbf{u}_k, \mathbf{u}_{k-1}) \\ \nonumber
    \text{s.t.} \quad \boldsymbol{\zeta}^s_{k+1} & = f^s(\boldsymbol{\zeta}^s_k, \boldsymbol{u}_k), \quad  \boldsymbol{\zeta}^s_{k} \models \phi \quad \forall \phi \in \boldsymbol{\Phi}_{active,k},  \\\nonumber
    & k = t,...,t+H-1
\end{align}
where $\textbf{u}^*_{t: \; t+H-1} = [\textbf{u}^*_{t}, ..., \textbf{u}^*_{t+H-1}]$ is the optimal control sequence over the planning horizon of the MPC. If the MPC problem gets infeasible in a particular time step, we consider the optimal solution from the last time step and pass it to the low-level controller.

\subsection{\textbf{Low-level Controller}}
In this level (Fig.~\ref{fig:lowchart}), we use the solution of the high-level controller to simulate the future trajectory of ego, over the planning horizon of the MPC. The main objective of this level is to make corrections to the performance of ego with respect to the cost function~\eqref{eqn:mpccost}, based on the detailed models that describe ego and environment. Runtime monitors are constructed from the active traffic constraints and applied to the vehicle's trajectory to verify correctness of its behavior. 


\subsubsection{Detailed Vehicle Model}
\label{section:detailedbicycle}
The detailed model used in this level is the four-wheel vehicle model, similar to the one used in \cite{turri2013linear}, with the nonlinear Pacejka model for computing the tire forces \cite{pacejka2005tire}, and with the same notation as the one used in the general bicycle model in~\eqref{eqn:simplebicycle}. We consider the state vector of ego in the detailed dynamic model at time step $t$ as $\boldsymbol{\zeta}^d_{t} = [X_t,Y_t,\dot{x}_t,\dot{y}_t,\psi_t,\dot{\psi}_t]^\top$, where $\dot{x}$ and $\dot{y}$ are the velocity components with respect to the local coordinate system in the car body frame and $\dot{\psi}$ is the rate of change of the heading angle. If we denote the control input by $\mathbf{u}_t=[\delta_t,\gamma_t]^\top$, then the discrete-time version of the detailed model can be described as $\boldsymbol{\zeta}^d_{t+1} = f^d(\boldsymbol{\zeta}^d_{t},\mathbf{u}_t)$.

\subsubsection{Detailed Environment Model}
\label{section:detailedenvironment}
In detailed model of environment, we consider the bounding boxes of vehicles for checking the collision avoidance constraints.  The axis-aligned bounding box of the nearby vehicle $i$ at time step $t$ is denoted by $\mathcal{BB}_{i,t}$, and we define $\mathcal{BB}_{near,t} = \bigcup_{i \in I_{near,t}} \mathcal{BB}_{i,t}$ as the union of bounding boxes of nearby vehicles. The bounding box of ego itself is denoted by $\mathcal{BB}_t$. We can verify the satisfaction of collision avoidance constraints by checking the intersection between the bounding boxes. Hence, in the low-level controller we consider the predicate $(collision\_avoidance)$ true at time step $t$ if $\mathcal{BB}_t \, \cap \, \mathcal{BB}_{near,t} = \oldemptyset$, where we use notation of Section~\ref{section:generalenvironmentmodel}.

\subsubsection{Runtime Monitoring}
We use runtime monitoring techniques from~\cite{ulus2019online} to verify the satisfaction of the active traffic constraints, in the detailed setting, i.e., complex ego dynamic model and detailed environment model as defined above.
We simulate the trajectory of the vehicle in the detailed framework over the MPC horizon, and then use the constructed monitors to verify the satisfaction of the traffic rules.
We first examine the optimal control $\mathbf{u}_{opt}$ from the high-level controller, and if it satisfies all constraints, it is applied on the vehicle; otherwise, we try near-optimal control inputs that are randomly sampled from the $r$-ball $C = \{ \mathbf{u} \mid \| \mathbf{u} -\mathbf{u}_\mathit{opt} \| \leq r \}$ around the optimal control in the control space, for a small positive value $r$.
We continue the above procedure until the simulated trajectory of the vehicle with a control from region $C$ satisfies all constraints. 

If the controller cannot find the best control input within the sampling threshold of the low-level, we apply full braking to the vehicle. Handling infeasibility is a big issue in the field of self-driving cars. We chose full-breaking, because we are interested in urban driving, where the most sensible action is to stop.  This is the recommended course of action by law in case a driver does not know what to do.

\section{Results}
\label{section:results}
The performance and capabilities of our proposed control method are validated through four urban-driving simulation scenarios. We have used CARLA \cite{dosovitskiy2017carla} in Python, which is an open-source autonomous driving simulator that uses highly-detailed models for the traffic participants and urban environments. We use Gurobi package \cite{optimization2014inc} for solving the MPC problem in the high-level controller. In all scenarios ego must follow a reference path as close as possible, but also satisfy a set of traffic rules, which we detail for each scenario. In the simple representation the scenarios, the reference path is shown by a red line that connects the initial point A to the goal point B. To measure the runtime performance of our control algorithm, we have used frequency of the system's clock, which represents the rate of the control loop and environment update. This quantity is expressed in Frame Per Second (FPS) in our figures. In all simulations we set the parameters as follow: $H=10, \, l_f = 2.11\, m, \,  r_{near} = 10\, m, \, r = 0.3\, m,\, D_{safe}=1\, m$, and time step $\Delta t = 0.1\,s$.

\subsubsection{\textbf{Comparison scenario}}
\begin{figure} [htb]
    \centering
    \subfigure[]
    {\includegraphics[width=0.40\columnwidth]{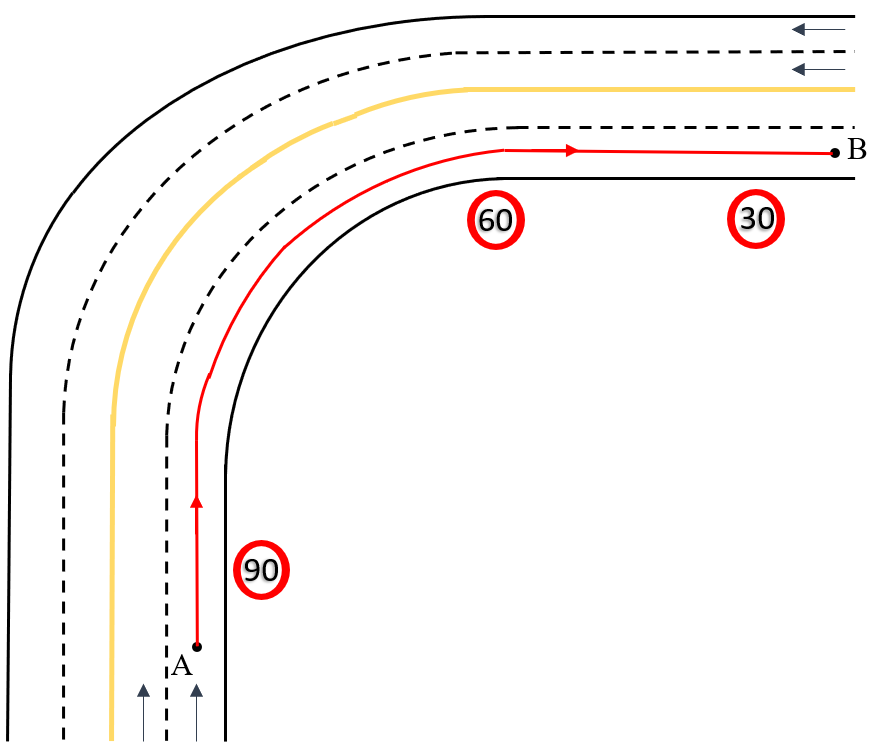}\label{fig:comparisonscenario}}
    \subfigure[]
    {\includegraphics[width=0.40\columnwidth]{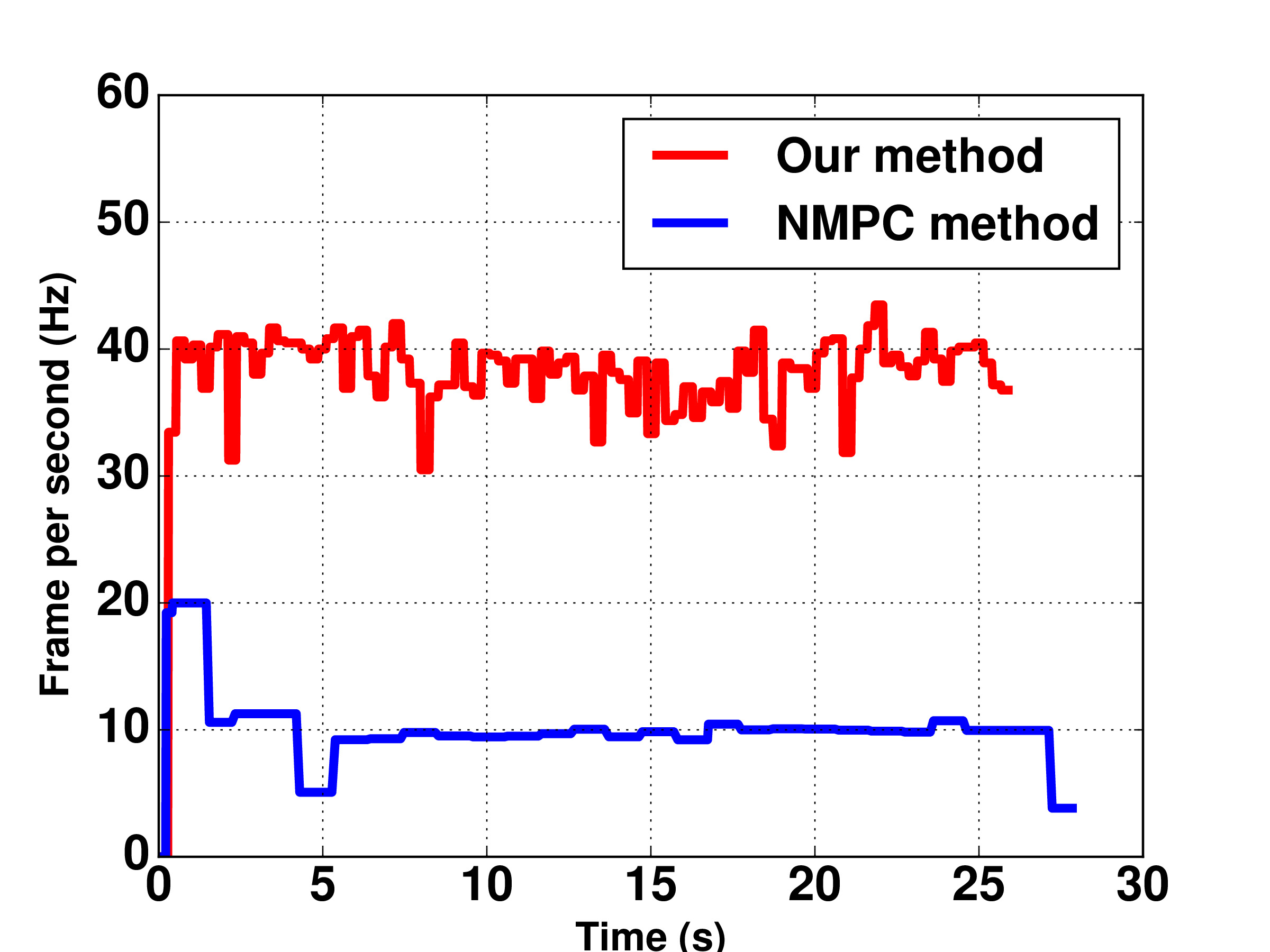}\label{fig:comparisonruntime}} 
    \caption{(a) Simple representation of the comparison scenario (b) FPS of our control method compared to the NMPC approach}
\end{figure}

To show the performance of our control algorithm compared to the existing works in the literature based on NMPC, we have designed a scenario where the vehicle has to follow a reference path in a curved road with different speed limit signs along it (Fig. \ref{fig:comparisonscenario}). 
We first formulated the NMPC problem with the detailed dynamic model used in the bottom-level of our algorithm, but the runtime performance of this method was less than $5 HZ$ on average and it led to instability of the controller. Then for having results that are comparable with our algorithm, we formulated the NMPC controller with a simple bicycle model, and here we present the figures for the second approach. Also we implemented our control approach and the NMPC method with the same planning horizon, same reference path, and the same cost function for the optimization problem. As shown in Fig.~\ref{fig:comparisonruntime}, our proposed method based on FPS, is much faster than the NMPC.

\begin{figure} [htb]
    \centering
    \vspace{-2mm}
    \subfigure[Our method's error]
    {\includegraphics[width=0.45\columnwidth]{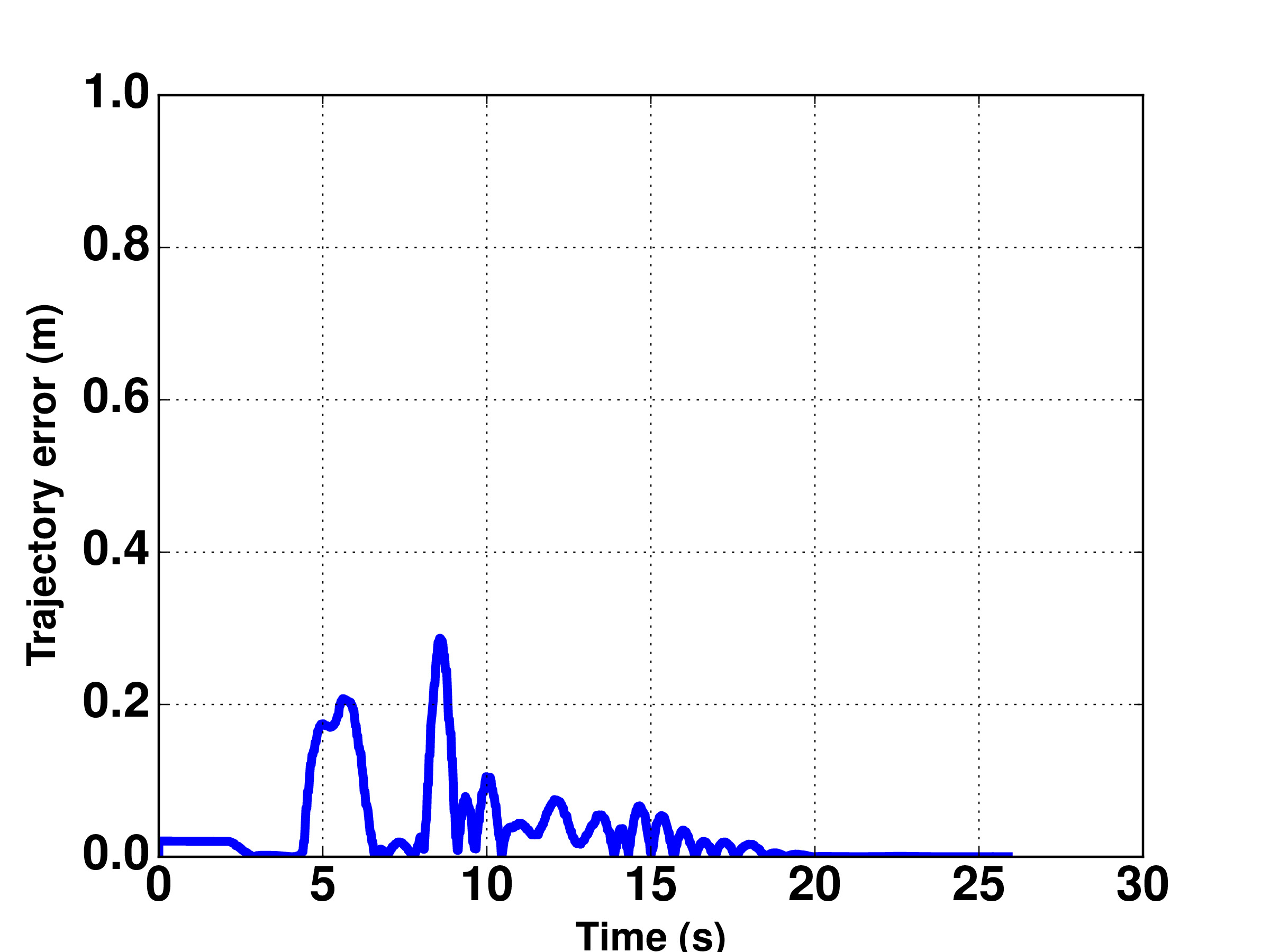}\label{fig:comparisonourerror}}
    \subfigure[NMPC method's error]
    {\includegraphics[width=0.45\columnwidth]{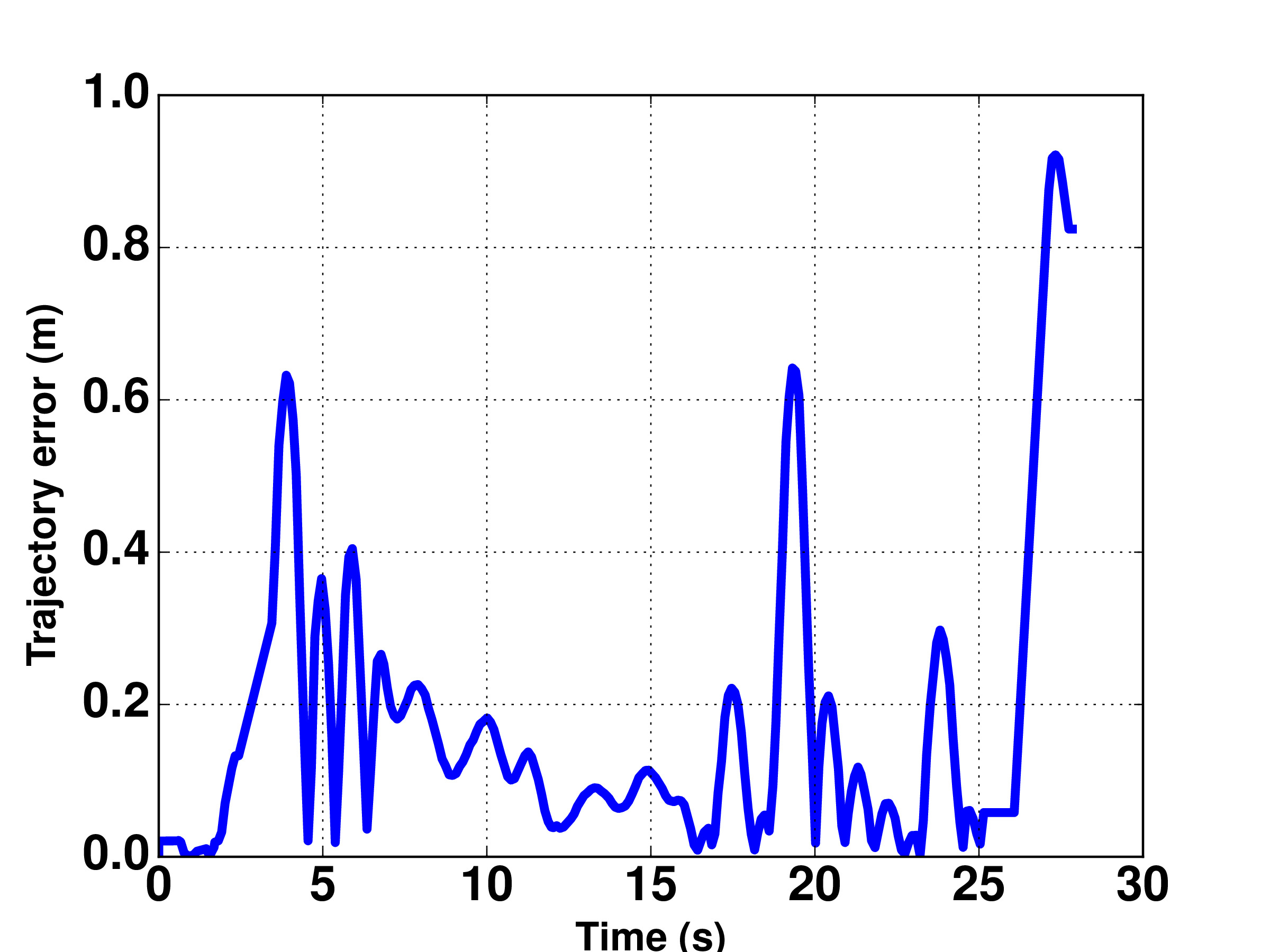}\label{fig:comparisonnmpcerror}} 
    \caption{Trajectory errors with respect to the reference path}
    \label{fig:comparisonerrors}
    \vspace{-2mm}
\end{figure}

\begin{figure}[htb]
    \centering
    \subfigure[Our method's speed profile]
    {\includegraphics[width=0.45\columnwidth]{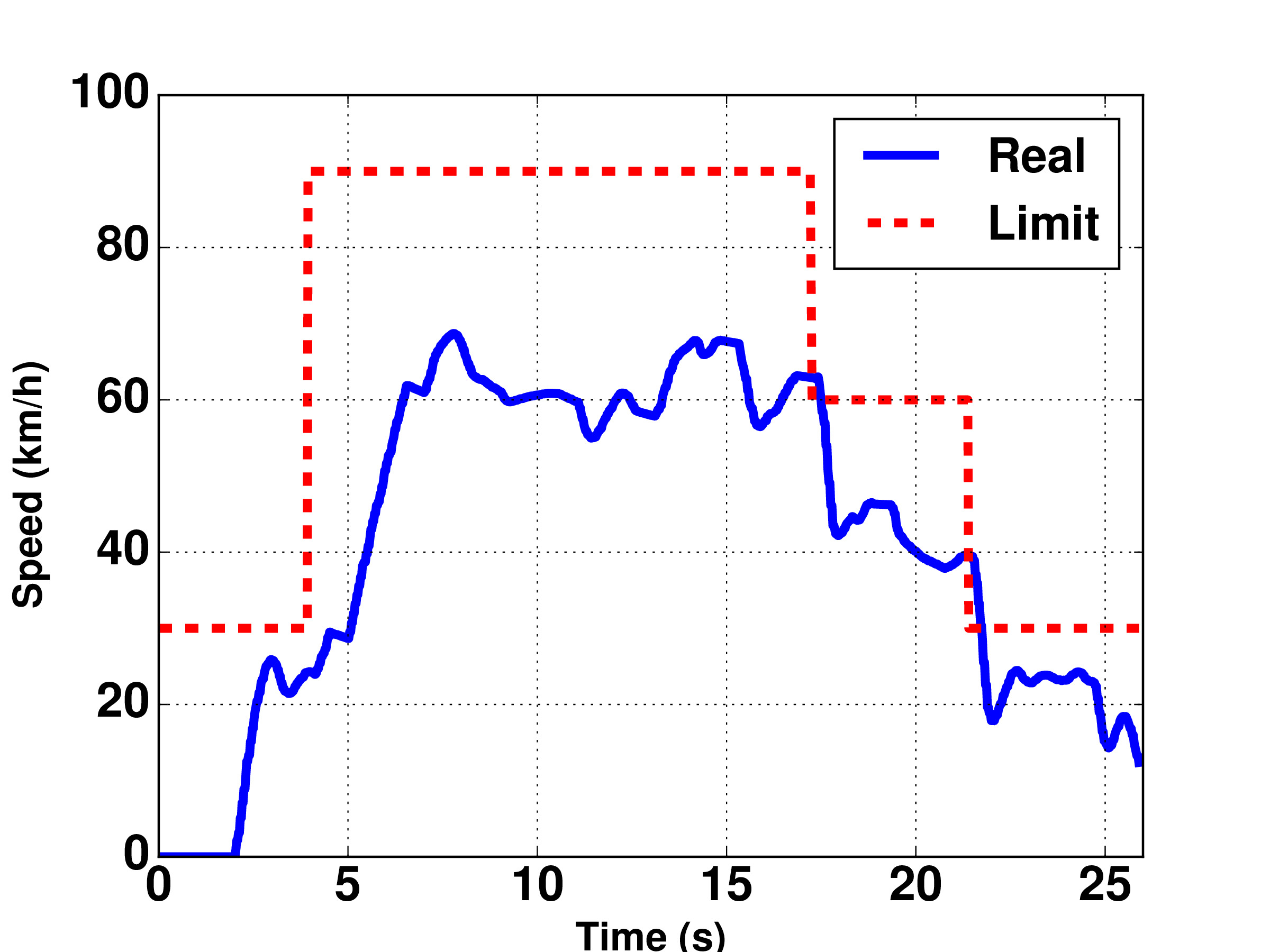}\label{fig:comparisonourvelocity}}
    \subfigure[NMPC method's speed profile]
    {\includegraphics[width=0.45\columnwidth]{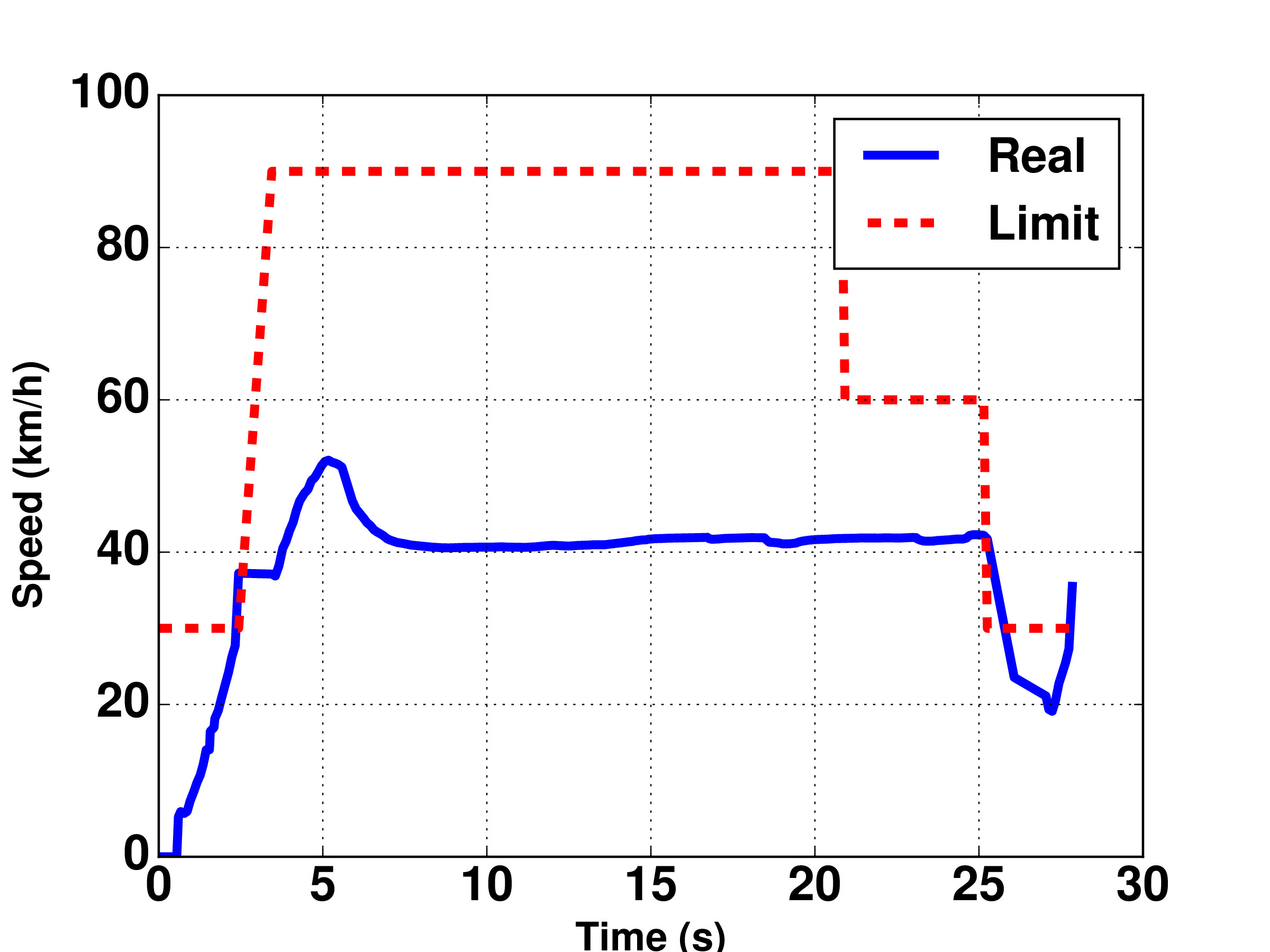}\label{fig:comparisonnmpcvelocity}} 
    \caption{Speed profiles of the algorithms. Our algorithm reaches higher legal speeds on average, compared to the NMPC implementation.}
    \vspace{-4.5mm}
    \label{fig:comparisonvelocities}
\end{figure}

In Fig. \ref{fig:comparisonerrors} the error of each algorithm's trajectory with respect to the reference path is plotted over time, which shows our algorithm follows the reference path with smaller error on average, compared to the NMPC method. Maximum tracking error of our controller is 0.29m, while this value for the NMPC implementation is 0.93m. Finally in Fig. \ref{fig:comparisonvelocities} the speed profiles of the algorithms are presented, where the maximum speed limits are shown by dash lines. Overall, we believe that the NMPC problem is more likely to get stuck in the locally-optimal solutions, than our quadratic programming problem with MILCs, and that can adversely affect the performance of the NMPC approach.

\subsubsection{\textbf{Traffic rules scenario}}
\begin{figure}[htb]
    \centering
    \subfigure[]
    {\includegraphics[width=0.45\columnwidth]{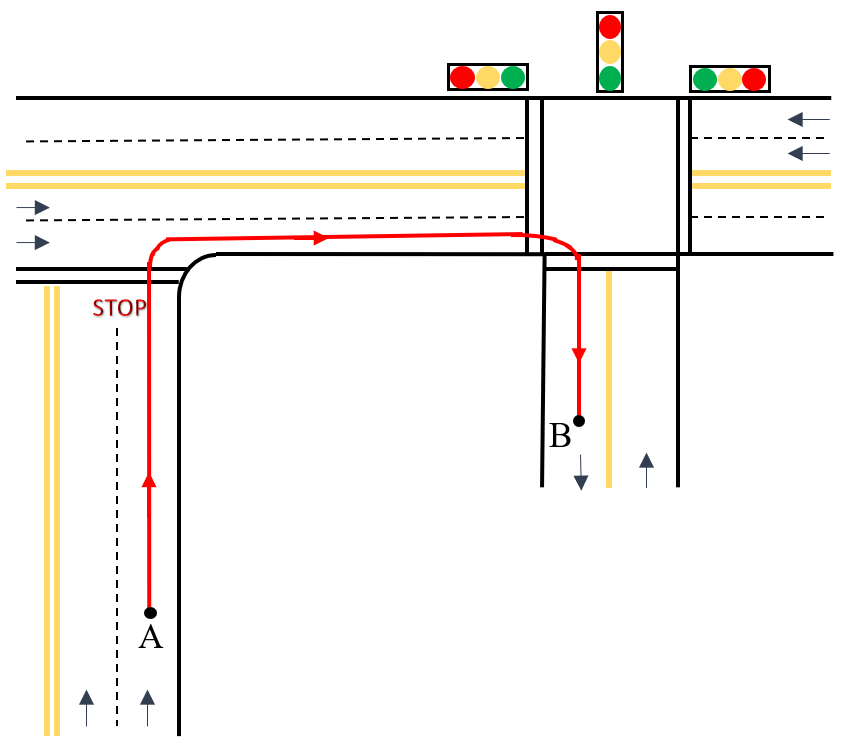}\label{fig:trafficrulescenario}}
    \subfigure[]
    {\includegraphics[width=0.45\columnwidth]{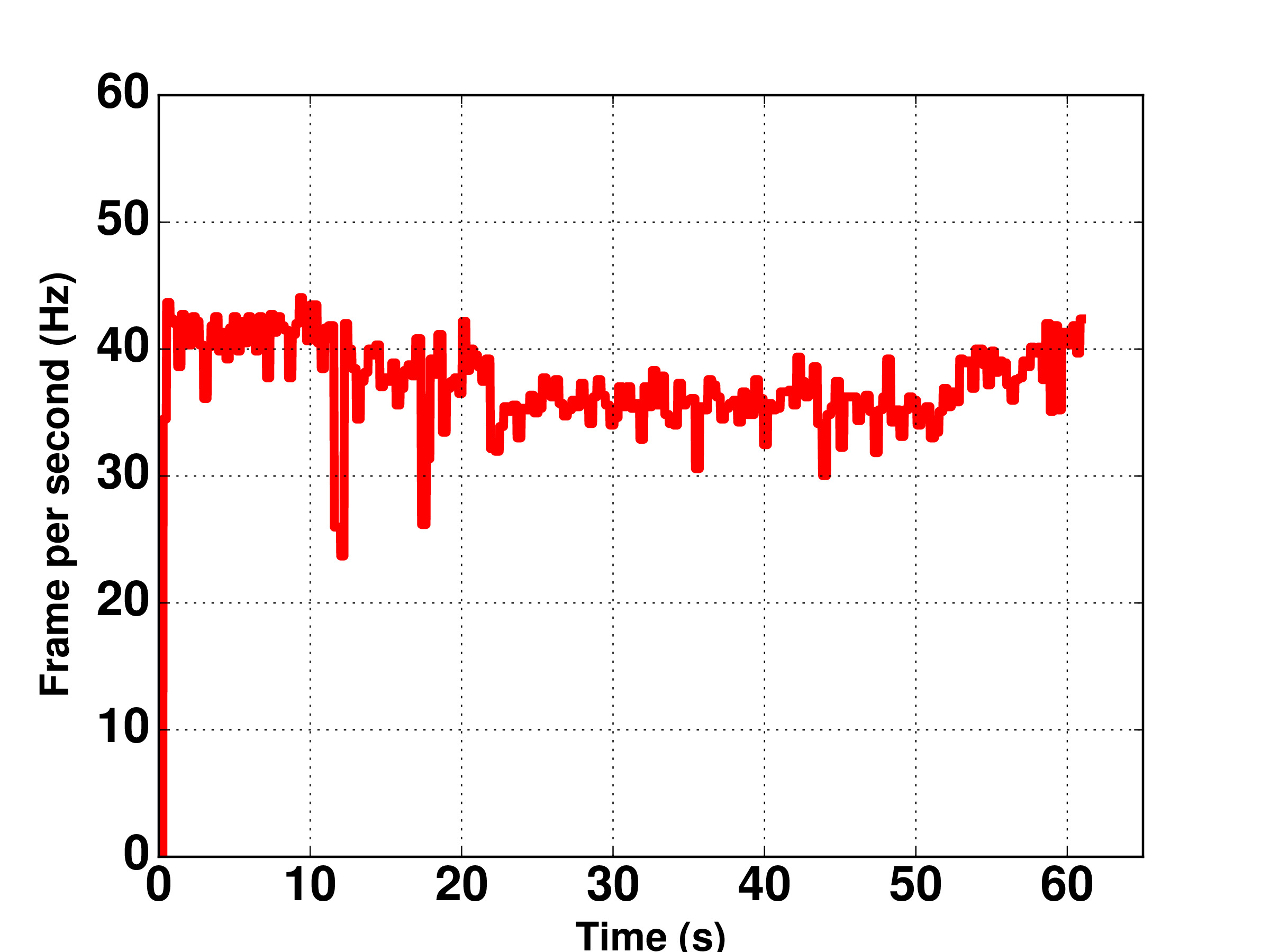}\label{fig:trafficrulesruntime}}
    \caption{(a) Simple representation of traffic rules scenario, (b) FPS of our controller.}
    \vspace{-3mm}
\end{figure}

This scenario shows how well our algorithm satisfies the traffic rules. In this scenario while the vehicle is following the reference path, it reaches a stop sign and a traffic light (Fig.~\ref{fig:trafficrulescenario}), and has to adjust its velocity to satisfy the rules. Fig. \ref{fig:trafficrulesruntime} shows that overall, our controller maintains a fast performance during its operation.  

\begin{figure}[htb]
    \centering
    \subfigure[]
    {\includegraphics[width=0.45\columnwidth]{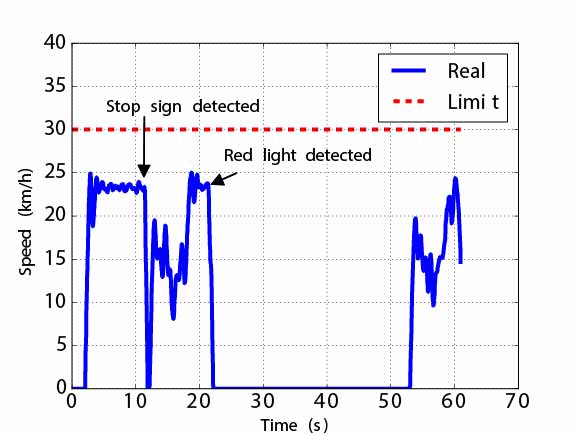} \label{fig:trafficrulesvelocity}} 
    \subfigure[]
    {\includegraphics[width=0.45\columnwidth]{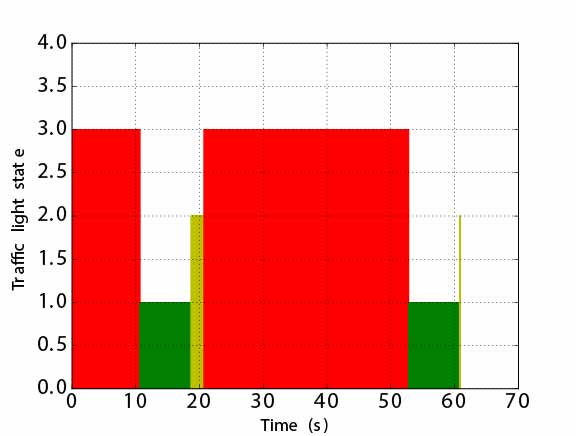} \label{fig:trafficruleslight}}
    \caption{(a) Speed profile of ego, and (b) traffic light's state over time}
    \vspace{-4mm}
\end{figure}

In Fig. \ref{fig:trafficrulesvelocity} the speed profile of the vehicle is presented together with the traffic light's state over time in Fig. \ref{fig:trafficruleslight} that shows the fast performance of ego with respect to the events happening in its environment. In Fig.~\ref{fig:trafficrulesvelocity}, when ego detects the stop sign at time 11.2s, it starts decreasing its velocity to stop near the stop sign at time 11.9s. Afterwards, it detects the red light at time 21.7s, and stops at time 22.1s. Its velocity is zero until it detects the green light at time 52.8s, when it starts moving again.

\subsubsection{\textbf{Safety scenario}}
\begin{figure}[h]
    \centering
    \subfigure[]
    {\includegraphics[width=0.45\columnwidth]{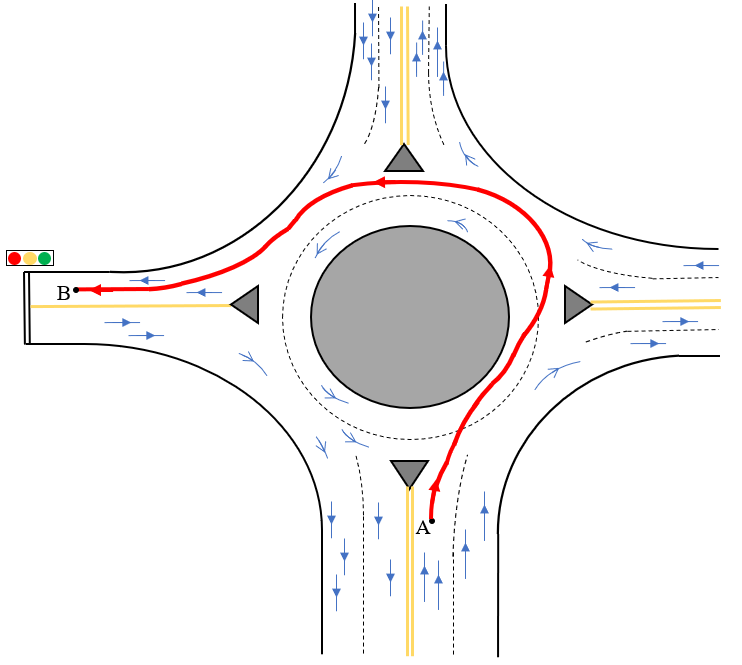}\label{fig:safetyscenario}}
    \subfigure[]
    {\includegraphics[width=0.45\columnwidth]{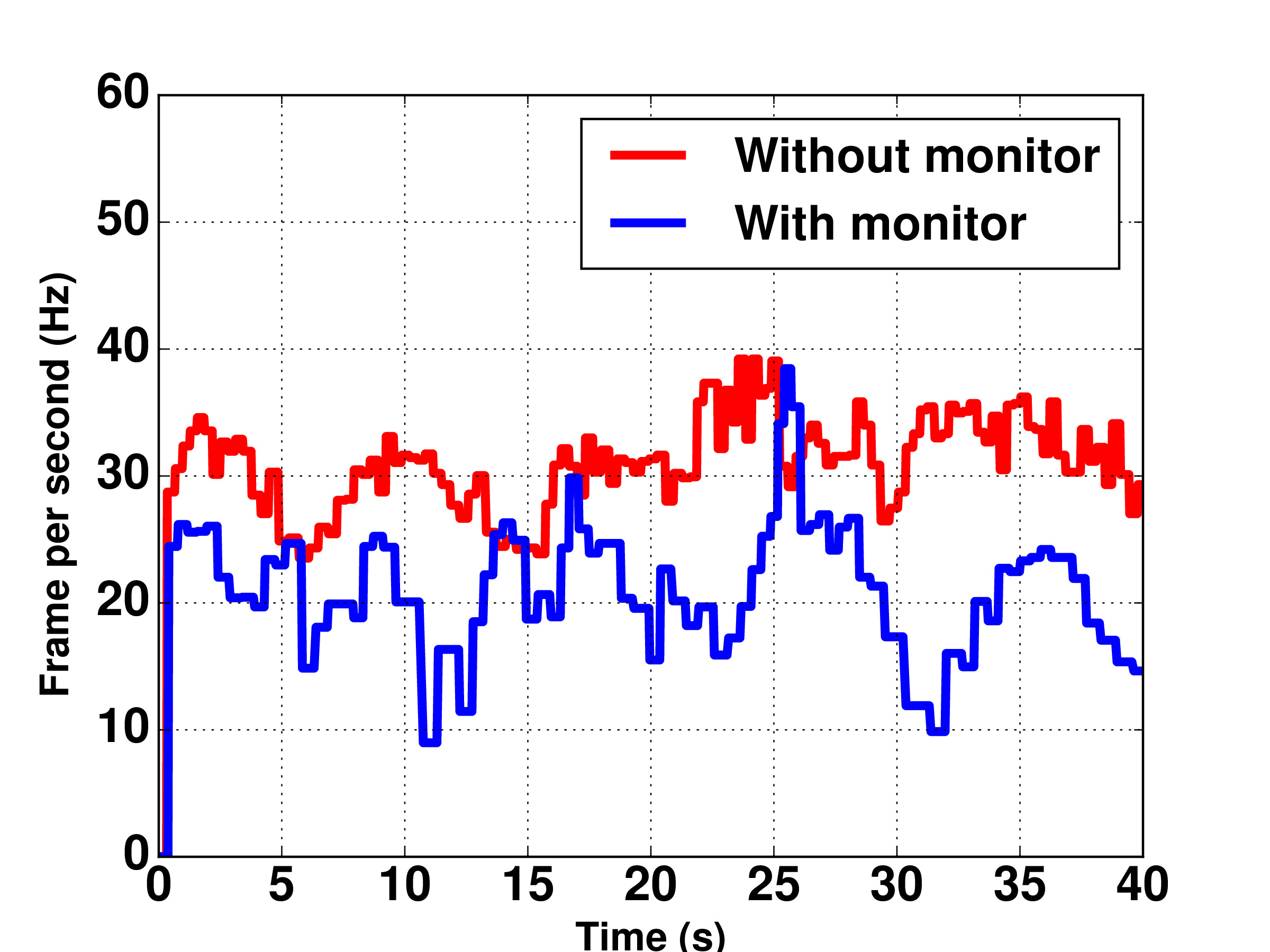}\label{fig:safetyruntime}}
    \caption{(a) Simple representation of safety scenario (b) FPS of the controller, with and without monitoring}
    \vspace{-2.5mm}
\end{figure}

This scenario shows how the low-level controller compensates for the mismatch between the real world and the simple models in the high-level controller, to ensure safety. In Fig.~\ref{fig:safetyscenario} a simple representation of the scenario is shown, where there are 160 other vehicles in the environment and the blue arrows show their direction of motion. In Fig.~\ref{fig:safetyruntime} the runtime speed of our controller, with and without applying the runtime monitoring, is represented. We see the average frame per second rate in the "with monitor" case is less than the "without monitor" one. This is reasonable, because low-level controller considers more complicated models, with more constraints, and in "with monitor" implementation it decreases runtime speed on average.

\begin{figure}[htb]
\centering
\vspace{-4mm}
\includegraphics[width=0.60\columnwidth]{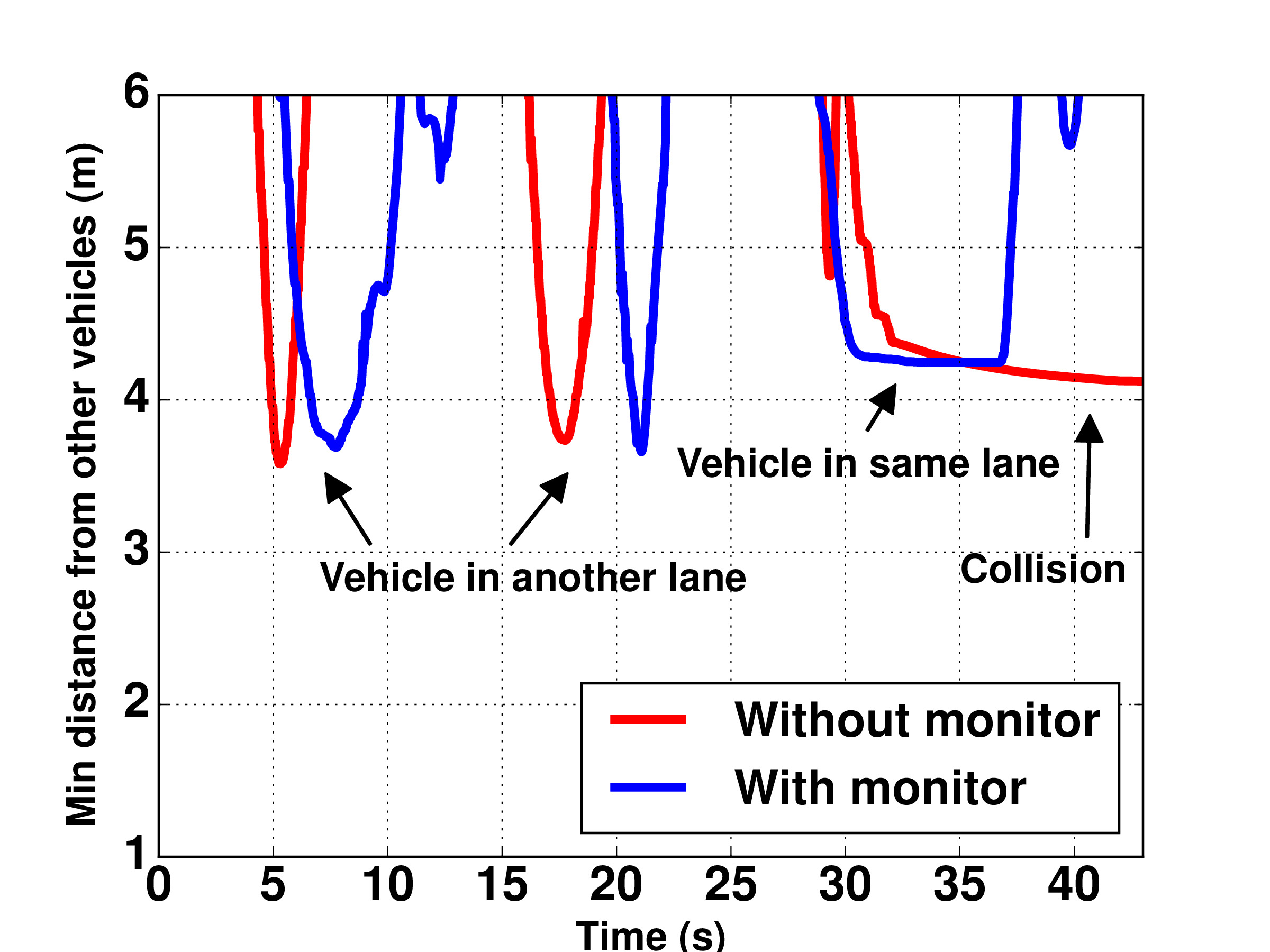}
\caption{Minimum distance between ego and its nearby vehicles over time.}
\vspace{-3mm}
\label{fig:safetymindistancewhole}
\end{figure}

The main purpose of this scenario is to show the role of low-level controller to ensure safety, and it is presented in Fig.~\ref{fig:safetymindistancewhole}. The minimum distance of ego from nearby vehicles is plotted over time in both cases, with and without monitoring. Bounding boxes of all vehicles have same length (4.22m) and width (1.8m); therefore, if two vehicles in the same (parallel) lane(s) get closer than 4.22m (1.8m), a collision has happened between them. In this scenario, in both of "with monitor" and "without monitor" cases and around the times $\sim$6s and $\sim$19s, ego detects its nearest nearby vehicle in the parallel lane, and their distance is bigger than 3.5m, which means no collision happens. Around the time 32s, ego detects another stationary vehicle in the same lane, and ego decreases speed to stop behind it. We see that in the "without monitor" case, the distance between the vehicles gradually gets less than 4.22m and eventually a collision occurs between them, while in the "with monitor" case ego maintains the distance bigger than 4.22 m. Hence, the low-level runtime monitoring helps to ensure safety for ego.

\subsubsection{\textbf{Scalability scenario}}
\label{section:scalabilityscenario}
\begin{figure}[htb]
    \centering
    \subfigure[]
    {\includegraphics[width=0.42\columnwidth]{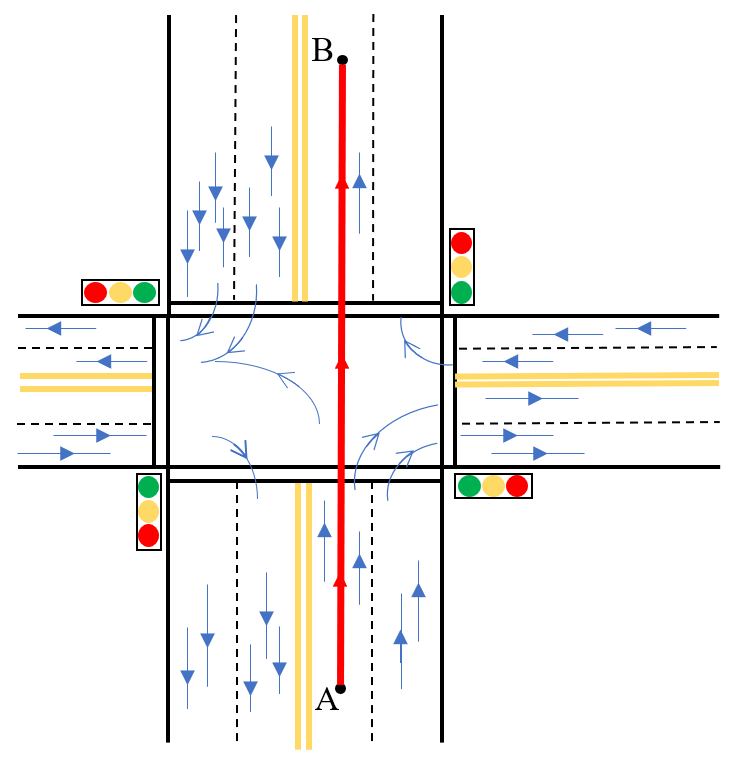}\label{fig:scalabilityscenario}}
    \subfigure[]
    {\includegraphics[width=0.45\columnwidth]{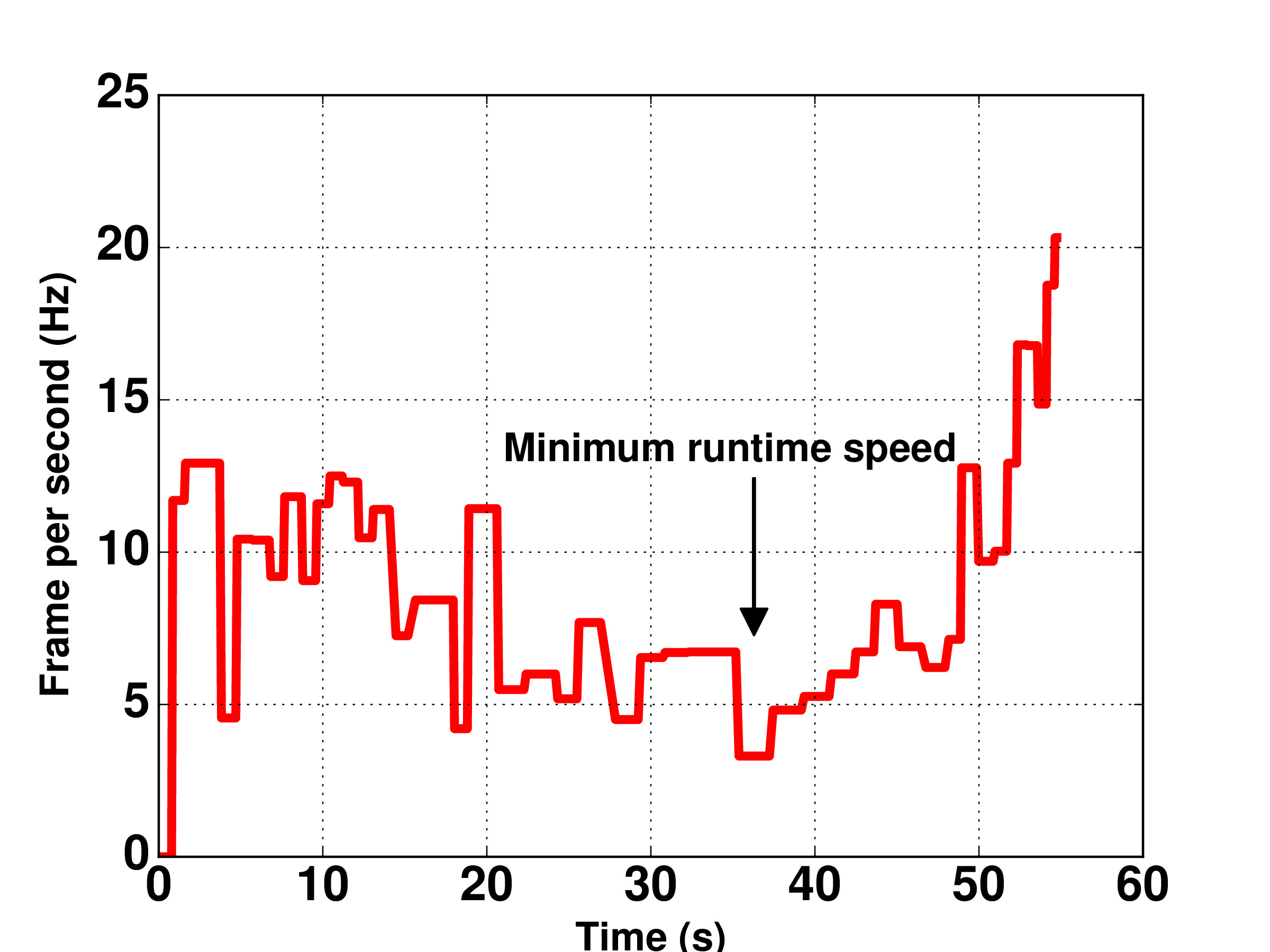}\label{fig:scalabilityruntime}}
    \subfigure[]
    {\includegraphics[width=0.45\columnwidth]{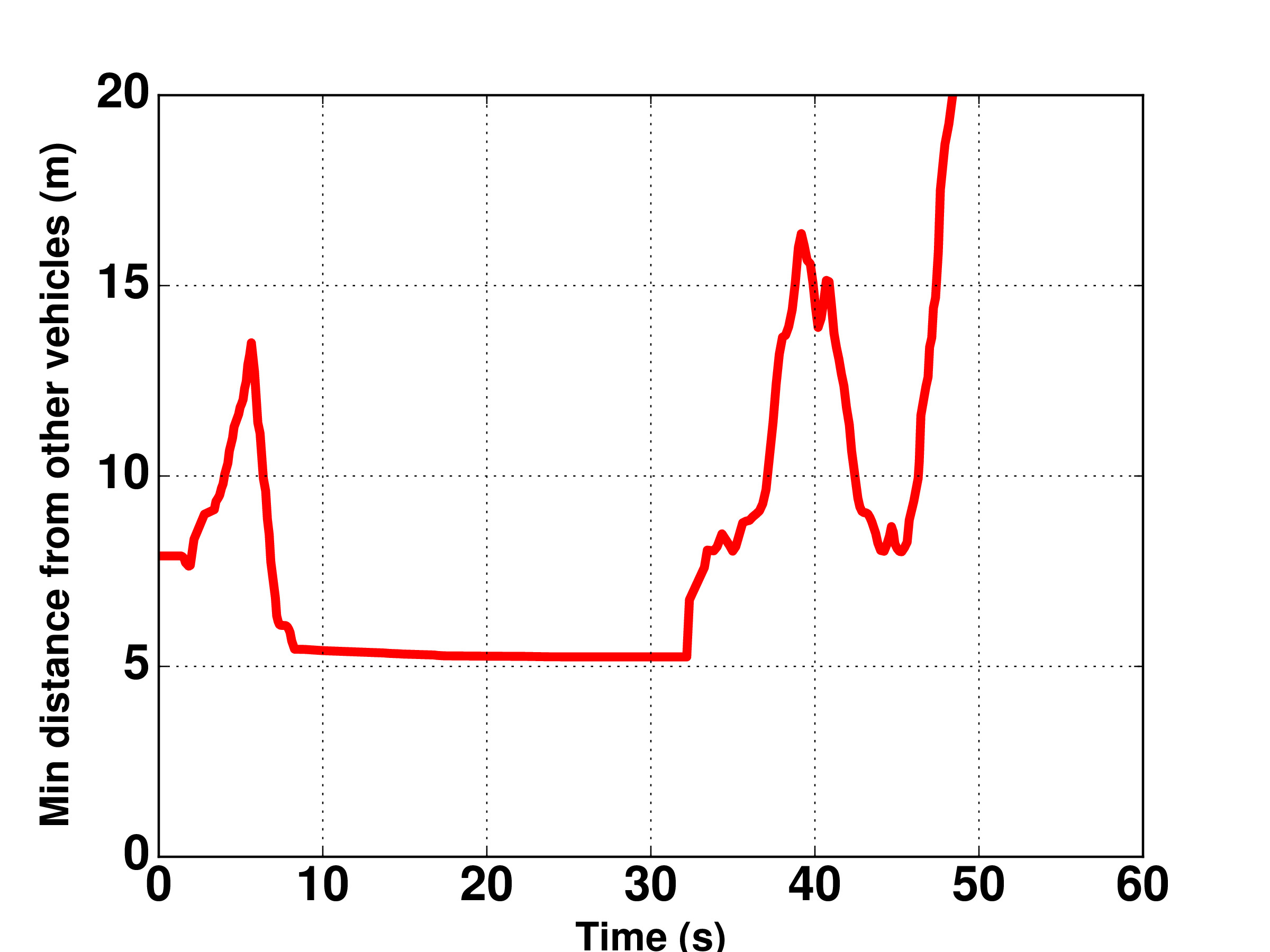}\label{fig:scalabilitymindistance}}
    \subfigure[]
    {\includegraphics[width=0.45\columnwidth]{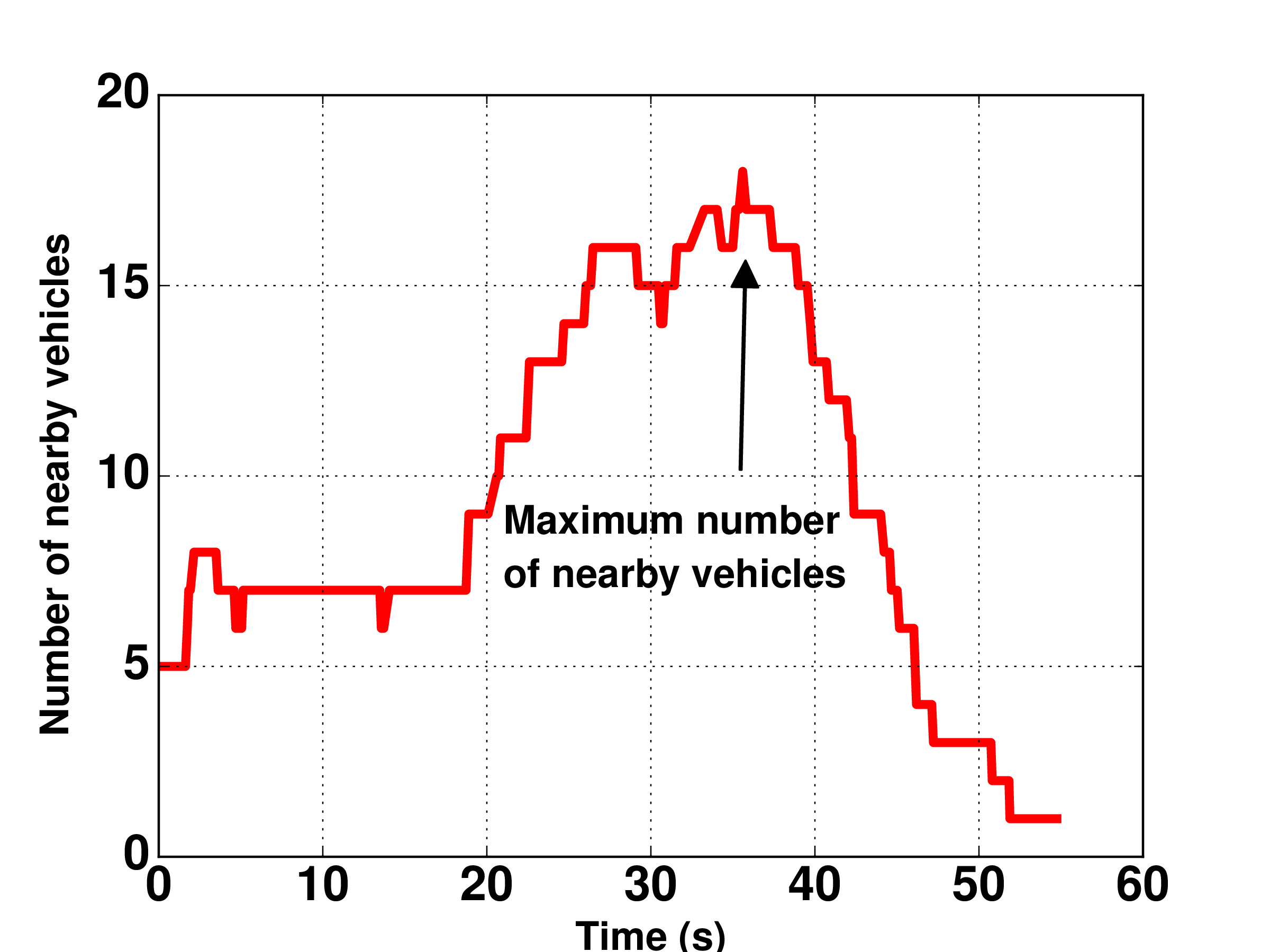}\label{fig:scalabilitynearveh}}
    \caption{(a) Simple representation of scalability scenario, (b) FPS of algorithm over time, (c) minimum distance from nearby vehicles, (d) number of nearby vehicles that have been detected}
    \vspace{-4.5mm}
\end{figure}

This scenario shows how our proposed control method scales with the number of nearby vehicles. In Fig. \ref{fig:scalabilityscenario} a simple representation of the scenario is provided, where there are 121 other vehicles in the environment and we set $r_{near} = 40$ m. In Fig. \ref{fig:scalabilityruntime} and \ref{fig:scalabilitynearveh}, the frame per second rate and the number of nearby vehicles that ego detects, are plotted over time. We see that during most of the runtime of the scenario, the vehicle detects on average more than 6 vehicles, and close to the time 36s, it detects 18 vehicles and predicts their future trajectories in a detailed framework, which drops the FPS to its lowest value of 4Hz. In Fig. \ref{fig:scalabilitymindistance}, the minimum distance of ego relative to its nearby vehicles is presented over time and it verifies that ego always keeps a safe distance from nearby vehicles.

\section{Conclusion and Future Work}
\label{section:conclusion}

This paper presents a two-level control approach for fully autonomous vehicles. In the high-level controller we formulate a MPC problem with MILCs, based on simple models for ego and the environment. Then, specification-based runtime monitoring techniques are used in the low-level controller to check the correctness of ego's trajectories against complex models. Our control is over both steering and throttle/braking of ego and we obtain substantial performance improvements, in terms of quality and runtime, over existing NMPC approaches.

\bibliographystyle{IEEEtran}
\bibliography{IEEEexample}

\begin{thebibliography}{10}
\providecommand{\url}[1]{#1}
\csname url@rmstyle\endcsname
\providecommand{\newblock}{\relax}
\providecommand{\bibinfo}[2]{#2}
\providecommand\BIBentrySTDinterwordspacing{\spaceskip=0pt\relax}
\providecommand\BIBentryALTinterwordstretchfactor{4}
\providecommand\BIBentryALTinterwordspacing{\spaceskip=\fontdimen2\font plus
\BIBentryALTinterwordstretchfactor\fontdimen3\font minus
  \fontdimen4\font\relax}
\providecommand\BIBforeignlanguage[2]{{%
\expandafter\ifx\csname l@#1\endcsname\relax
\typeout{** WARNING: IEEEtran.bst: No hyphenation pattern has been}%
\typeout{** loaded for the language `#1'. Using the pattern for}%
\typeout{** the default language instead.}%
\else
\language=\csname l@#1\endcsname
\fi
#2}}

\bibitem{vahidi2003research}
A.~Vahidi and A.~Eskandarian, ``Research advances in intelligent collision
  avoidance and adaptive cruise control,'' \emph{IEEE transactions on
  intelligent transportation systems}, vol.~4, no.~3, pp. 143--153, 2003.

\bibitem{liu2007development}
J.-F. Liu, J.-H. Wu, and Y.-F. Su, ``Development of an interactive lane keeping
  control system for vehicle,'' in \emph{2007 IEEE Vehicle Power and Propulsion
  Conference}, pp. 702--706.

\bibitem{camacho2013model}
E.~F. Camacho and C.~B. Alba, \emph{Model predictive control}.\hskip 1em plus
  0.5em minus 0.4em\relax Springer Science \& Business Media, 2013.

\bibitem{zhang2016model}
R.~Zhang, F.~Rossi, and M.~Pavone, ``Model predictive control of autonomous
  mobility-on-demand systems,'' in \emph{2016 IEEE International Conference on
  Robotics and Automation}, pp. 1382--1389.

\bibitem{anderson2010optimal}
S.~J. Anderson, S.~C. Peters, T.~E. Pilutti, and K.~Iagnemma, ``An
  optimal-control-based framework for trajectory planning, threat assessment,
  and semi-autonomous control of passenger vehicles in hazard avoidance
  scenarios,'' \emph{International Journal of Vehicle Autonomous Systems},
  vol.~8, no. 2-4, pp. 190--216, 2010.

\bibitem{faulwasser2009model}
T.~Faulwasser, B.~Kern, and R.~Findeisen, ``Model predictive path-following for
  constrained nonlinear systems,'' in \emph{Proceedings of the 48h IEEE
  Conference on Decision and Control held jointly with 2009 28th Chinese
  Control Conference}, pp. 8642--8647.

\bibitem{weiskircher2017predictive}
T.~Weiskircher, Q.~Wang, and B.~Ayalew, ``Predictive guidance and control
  framework for (semi-) autonomous vehicles in public traffic,'' \emph{IEEE
  Transactions on control systems technology}, vol.~25, no.~6, pp. 2034--2046,
  2017.

\bibitem{schwarting2017parallel}
W.~Schwarting, J.~Alonso-Mora, L.~Pauli, S.~Karaman, and D.~Rus, ``Parallel
  autonomy in automated vehicles: Safe motion generation with minimal
  intervention,'' in \emph{2017 IEEE International Conference on Robotics and
  Automation}, pp. 1928--1935.

\bibitem{erlien2015shared}
S.~M. Erlien, S.~Fujita, and J.~C. Gerdes, ``Shared steering control using safe
  envelopes for obstacle avoidance and vehicle stability,'' \emph{IEEE
  Transactions on Intelligent Transportation Systems}, vol.~17, no.~2, pp.
  441--451, 2015.

\bibitem{gray2013robust}
A.~Gray, Y.~Gao, J.~K. Hedrick, and F.~Borrelli, ``Robust predictive control
  for semi-autonomous vehicles with an uncertain driver model,'' in \emph{2013
  IEEE Intelligent Vehicles Symposium}, pp. 208--213.

\bibitem{liu2015stochastic}
C.~Liu, A.~Carvalho, G.~Schildbach, and J.~K. Hedrick, ``Stochastic predictive
  control for lane keeping assistance systems using a linear time-varying
  model,'' in \emph{2015 American Control Conference}.\hskip 1em plus 0.5em
  minus 0.4em\relax IEEE, pp. 3355--3360.

\bibitem{turri2013linear}
V.~Turri, A.~Carvalho, H.~E. Tseng, K.~H. Johansson, and F.~Borrelli, ``Linear
  model predictive control for lane keeping and obstacle avoidance on low
  curvature roads,'' in \emph{16th international IEEE conference on intelligent
  transportation systems}, 2013, pp. 378--383.

\bibitem{sadigh2016planning}
D.~Sadigh, S.~Sastry, S.~A. Seshia, and A.~D. Dragan, ``Planning for autonomous
  cars that leverage effects on human actions.'' in \emph{Robotics: Science and
  Systems}, vol.~2.\hskip 1em plus 0.5em minus 0.4em\relax Ann Arbor, MI, USA,
  2016.

\bibitem{schwarting2019social}
W.~Schwarting, A.~Pierson, J.~Alonso-Mora, S.~Karaman, and D.~Rus, ``Social
  behavior for autonomous vehicles,'' \emph{Proceedings of the National Academy
  of Sciences}, vol. 116, no.~50, pp. 24\,972--24\,978, 2019.

\bibitem{karlsson2018multi}
J.~Karlsson, C.-I. Vasile, J.~Tumova, S.~Karaman, and D.~Rus, ``Multi-vehicle
  motion planning for social optimal mobility-on-demand,'' in \emph{2018 IEEE
  International Conference on Robotics and Automation}, pp. 7298--7305.

\bibitem{tumova2013least}
J.~Tumova, G.~C. Hall, S.~Karaman, E.~Frazzoli, and D.~Rus, ``Least-violating
  control strategy synthesis with safety rules,'' in \emph{Proceedings of the
  16th international conference on Hybrid systems: computation and control},
  2013, pp. 1--10.

\bibitem{vasile2017minimum}
C.-I. Vasile, J.~Tumova, S.~Karaman, C.~Belta, and D.~Rus, ``Minimum-violation
  scltl motion planning for mobility-on-demand,'' in \emph{2017 IEEE
  International Conference on Robotics and Automation}, pp. 1481--1488.

\bibitem{censi2019liability}
A.~Censi, K.~Slutsky, T.~Wongpiromsarn, D.~Yershov, S.~Pendleton, J.~Fu, and
  E.~Frazzoli, ``Liability, ethics, and culture-aware behavior specification
  using rulebooks,'' in \emph{2019 International Conference on Robotics and
  Automation}.\hskip 1em plus 0.5em minus 0.4em\relax IEEE, pp. 8536--8542.

\bibitem{baier2008principles}
C.~Baier and J.-P. Katoen, \emph{Principles of model checking}.\hskip 1em plus
  0.5em minus 0.4em\relax MIT press, 2008.

\bibitem{ding2014ltl}
X.~Ding, M.~Lazar, and C.~Belta, ``Ltl receding horizon control for finite
  deterministic systems,'' \emph{Automatica}, vol.~50, no.~2, pp. 399--408,
  2014.

\bibitem{wongpiromsarn2012receding}
T.~Wongpiromsarn, U.~Topcu, and R.~M. Murray, ``Receding horizon temporal logic
  planning,'' \emph{IEEE Transactions on Automatic Control}, vol.~57, no.~11,
  pp. 2817--2830, 2012.

\bibitem{raman2014model}
V.~Raman, A.~Donz{\'e}, M.~Maasoumy, R.~M. Murray, A.~Sangiovanni-Vincentelli,
  and S.~A. Seshia, ``Model predictive control with signal temporal logic
  specifications,'' in \emph{53rd IEEE Conference on Decision and Control},
  2014, pp. 81--87.

\bibitem{sadraddini2015robust}
S.~Sadraddini and C.~Belta, ``Robust temporal logic model predictive control,''
  in \emph{2015 53rd Annual Allerton Conference on Communication, Control, and
  Computing (Allerton)}.\hskip 1em plus 0.5em minus 0.4em\relax IEEE, pp.
  772--779.

\bibitem{donze2010robust}
A.~Donz{\'e} and O.~Maler, ``Robust satisfaction of temporal logic over
  real-valued signals,'' in \emph{International Conference on Formal Modeling
  and Analysis of Timed Systems}.\hskip 1em plus 0.5em minus 0.4em\relax
  Springer, 2010, pp. 92--106.

\bibitem{bartocci2018specification}
E.~Bartocci, J.~Deshmukh, A.~Donz{\'e}, G.~Fainekos, O.~Maler,
  D.~Ni{\v{c}}kovi{\'c}, and S.~Sankaranarayanan, ``Specification-based
  monitoring of cyber-physical systems: a survey on theory, tools and
  applications,'' in \emph{Lectures on Runtime Verification}, 2018, pp.
  135--175.

\bibitem{deshmukh2017robust}
J.~V. Deshmukh, A.~Donz{\'e}, S.~Ghosh, X.~Jin, G.~Juniwal, and S.~A. Seshia,
  ``Robust online monitoring of signal temporal logic,'' \emph{Formal Methods
  in System Design}, vol.~51, no.~1, pp. 5--30, 2017.

\bibitem{maler2004monitoring}
O.~Maler and D.~Nickovic, ``Monitoring temporal properties of continuous
  signals,'' in \emph{Formal Techniques, Modelling and Analysis of Timed and
  Fault-Tolerant Systems}.\hskip 1em plus 0.5em minus 0.4em\relax Springer,
  2004, pp. 152--166.

\bibitem{baillieul2007control}
J.~Baillieul and P.~J. Antsaklis, ``Control and communication challenges in
  networked real-time systems,'' \emph{Proceedings of the IEEE}, vol.~95,
  no.~1, pp. 9--28, 2007.

\bibitem{pacejka2005tire}
H.~Pacejka, \emph{Tire and vehicle dynamics}.\hskip 1em plus 0.5em minus
  0.4em\relax Elsevier, 2005.

\bibitem{ulus2019online}
D.~Ulus, ``Online monitoring of metric temporal logic using sequential
  networks,'' \emph{arXiv preprint arXiv:1901.00175}, 2019.

\bibitem{dosovitskiy2017carla}
A.~Dosovitskiy, G.~Ros, F.~Codevilla, A.~Lopez, and V.~Koltun, ``Carla: An open
  urban driving simulator,'' \emph{arXiv preprint:1711.03938}, 2017.

\bibitem{optimization2014inc}
\BIBentryALTinterwordspacing
L.~Gurobi~Optimization, ``Gurobi optimizer reference manual,'' 2020. [Online].
  Available: \url{http://www.gurobi.com}
\BIBentrySTDinterwordspacing

\end{thebibliography}

\end{document}